\documentclass[lettersize,journal]{IEEEtran}
\usepackage{amsmath,amsfonts}
\usepackage{algorithmic}
\usepackage{algorithm}
\usepackage{array}
\usepackage[caption=false,font=normalsize,labelfont=sf,textfont=sf]{subfig}
\usepackage{textcomp}
\usepackage{stfloats}
\usepackage{url}
\usepackage{verbatim}
\usepackage{graphicx}
\usepackage{multirow}  
\usepackage{enumitem}  
\usepackage{cite}
\begin{document}

\title{WCL-BBCD: A Contrastive Learning and Knowledge Graph Approach to Named Entity Recognition}

\author{Renjie Zhou, Qiang Hu, Jian Wan, Jilin Zhang, Qiang Liu, Tianxiang Hu, Jianjun Li}



\maketitle

\begin{abstract}
Named Entity Recognition task is one of the core tasks of information extraction. Word ambiguity and word abbreviation are important reasons for the low recognition rate of named entities. In this paper, we propose a novel named entity recognition model WCL-BBCD (Word Contrastive Learning with BERT-BiLSTM-CRF-DBpedia), which incorporates the idea of contrastive learning. The model first trains the sentence pairs in the text, calculate similarity between sentence pairs, and fine-tunes BERT used for the named entity recognition task according to the similarity, so as to alleviate word ambiguity. Then, the fine-tuned BERT is combined with BiLSTM-CRF to perform the named entity recognition task. Finally, the recognition results are corrected in combination with prior knowledge such as knowledge graphs, so as to alleviate the low-recognition-rate problem caused by word abbreviations. The results of experimentals conducted on the CoNLL-2003 English dataset and OntoNotes V5 English dataset show that our model outperforms other similar models on.
\end{abstract}


\section{Introduction}
\IEEEPARstart{I}{nformation} extraction aims at extracting structured data from unstructured text. Information extraction is applied in many fields. For example, information extraction can be applied to recommender systems \cite{zhou2021hybrid} and web data retrieval systems \cite{he2011using}. Government departments use information extraction to understand public opinion \cite{conlon2015terrorism}, \cite{atkinson2009automated}. In the field of biomedicine, information extraction can be used to identify diseases caused by interactions between biomolecules \cite{vanegas2015overview}. Named entity recognition is an important task in the field of information extraction\cite{giorgi2019end}. Named entity recognition aims to identify named entities from texts of predefined semantic types, such as person, place, organization, et al. In this paper, we propose a novel model WCL-BBCD (\textbf{W}ord \textbf{C}ontrastive \textbf{L}earning with \textbf{B}ERT-\textbf{B}iLSTM-\textbf{C}RF-\textbf{D}Bpedia) based on contrastive learning and knowledge graph for named entity recognition, and demonstrate the effectiveness of contrastive learning in word embedding. 

The method based on pattern matching is the earliest method used for named entity recognition, with the basic step of having a domain-specific expert or academic construct a recognition template based on the features of a sample dataset. Although this method has a high recognition rate in a specific domain, it suffers from a problem of poor portability. As a result, researchers have gradually shifted their focus to the study of machine learning methods. Traditional machine learning methods treat named entity recognition as a sequential annotation task, that is, for each word in the text, a series of candidate entity labels are set, and then the probability of each word belonging to each entity label is predicted by machine learning models, and the entity label with the highest probability value is usually taken as the entity label of the word. Common traditional machine learning classification models used for named entity recognition include Support Vector Machines (SVM)\cite{li2004svm}, Decision Tree (DT) \cite{schapire2013explaining}, etc.

In recent years, with the rise of deep learning, large-scale pre-trained language representation models such as BERT\cite{devlin2019bert}, RoBERTa\cite{liu2019roberta}, Albert\cite{lan2019albert}, Xlnet\cite{yang2019xlnet}, etc. have achieved a dominant position in various natural language processing tasks (text generation, reading comprehension, text classification, POS tagging, entity recognition). The architectures of these models are mostly based on Transformer\cite{vaswani2017attention} which uses a self-attention mechanism to capture long-range dependencies between tokens. Transformer encoders and decoders complete self-supervised learning tasks such as predicting concealed tokens and predicting future tokens by pre-training on a large-scale text corpus. 

However, problems like word ambiguity and low recognition of abbreviated words persist. The problem of word ambiguity means that the same word has different meanings in different contexts. If the word is an entity, it may belong to different entity categories in different contexts. Especially for the names of people and organizations, the problem of misidentification between place names and organization names are more obvious. For example, the word ``Blackburn'' in ``Peter Blackburn'' means the name of a person, but in the sentence ``Blackburn announced on Wednesday they and Dalglish had parted by mutual consent.'' means the team. The abbreviation of the word is a common expression in English text. For example, the abbreviation of ``National Basketball Association'' is ``NBA''. People can well understand that ``NBA'' represents the American professional basketball league. However, when the computer trains the model without prior knowledge, it will be difficult for the models to understand the correct interpretation of abbreviated words, which causes the problem that existing models have low recognition rates for abbreviated entity words.

To alleviate the low-recognition-rate problem caused by word ambiguity, we incorporate the idea of contrastive learning\cite{he2020momentum}, \cite{chen2020simple} into the model. Through contrastive learning, the learned features of a sample can be distinguished from other samples. As a result, through contrastive learning, the entity types that ambiguous words belong to in different contexts can be distinguished as much as possible. For example, contrastive learning will make the word ``Blackburn'' (entity type is Person) in ``Peter Blackburn'' as dissimilar as possible to the word ``Blackburn'' (entity type is Organization) in sentence ``Blackburn announced on Wednesday they and Dalglish had parted by mutual consent.'', thereby improving the recognition accuracy of ambiguous entity word. We added a training process combining the idea of contrastive learning to fine-tune the pretrained language representation model BERT, and thus improve BRET's performance of encoding ambiguous words into better word embedding. The fine-tuning process of BERT is completed by the WCL (\textbf{W}ord \textbf{C}ontrastive \textbf{L}earning) component, and then the fine-tuned BERT is integrated with BBC (\textbf{B}ERT-\textbf{B}iLSTM-\textbf{C}RF) component to form the WCL-BBC (\textbf{W}ord \textbf{C}ontrastive \textbf{L}earning with \textbf{B}ERT-\textbf{B}iLSTM-\textbf{C}RF) model for named entity recognition.

For the second problem, BERT may split abbreviated words into meaningless words during encoding, resulting in a low recognition rate of abbreviated entity words. To alleviate the problem, we add DBpedia to WCL-BBC and finally form the WCL-BBCD model. DBpedia contains prior knowledge about entitiesm, which means, if an abbreviated word is an entity, the public knowledge graph may generally contain prior knowledge about the abbreviated entity word. We modify the recognition results of abbreviated words by utilizing the prior knowledge about abbreviations contained in the public knowledge graph, thereby effectively alleviating the problem of low-recognition-rate caused by abbreviated words, to improve the accuracy of the model. The experimental results show that WCL-BBCD achieves a higher F1-score on the CoNLL-2003\cite{sang2003introduction} and OntoNotes V5\cite{pradhan2013towards} English test sets.

The main contributions of this paper are summarized as follows:
\begin{enumerate}
    \item We propose WCL-BBC combined with the idea of contrastive learning to train a large number of positive and negative sample sentence pairs to fine-tune the word embedding encoding model BERT in BERT-BiLSTM-CRF, which effectively alleviates the problem of incorrect entity type recognition caused by word ambiguity.
    \item We added DBpedia to the WCL-BBC and finally form the WCL-BBCD model. DBpedia public knowledge graph is used to obtain the set of potential entities of the text corpus input sentences in WCL-BBC and filter out the potential entities present in DBpedia. Matching and correcting the entity types output by WCL-BBC with those of potential entities, effectively alleviating the problem of recognition errors caused by word abbreviations.
    \item WCL-BBCD alleviates the problem of word ambiguity and the low-recognition-rate of abbreviated words to a certain extent. WCL-BBCD has achieved a higher F1-score on the English test sets of CoNLL-2003 and OntoNotes V5 compared to other models cited in this paper.
\end{enumerate}

\section{Related Work}

\subsection{Named Entity Recognition}
The main task of named entity recognition is to identify and classify proper nouns such as persons and places, or quantitative phrases such as time, and date in the text. There have been many conferences or projects in history that have promoted the development of named entity recognition. The famous ones are MUC (Message Understanding Conference), CoNLL (Conference on Computational Natural Language Learning), and ACE project (Automatic Content Extraction).

In NER, deep learning-based model architectures have also gone through a process from word architecture, character architecture, words and character combination architecture to a combination of attention mechanism architecture. The word-based architecture inputs  word embedding for each single word in the sentence to RNN, typically using BiLSTM combined with CRF (BiLSTM-CRF) for NER as proposed by Jin\cite{jin2018improving}. The character-based architecture treats a sentence as a sequence of characters, and this sequence of characters is predicted by RNN or LSTM for each character label, and then the character labels are processed into word labels. The architectural based on the combination of words and characters concatenates the word embedding and the character embedding into LSTM to realize named entity recognition, typically using CNN-BiLSTM-CRF proposed by Ma\cite{ma2016end}.

According to the different datasets, the currently named entity recognition tagging methods mainly include IOB, BIO, BIOES, Markup, etc. In IOB tagging method, all entities start with ``I'' or ``B'' tags. ``B'' is only used to distinguish the boundary between two consecutive entities of the same type, not the starting position of the entity. When the word is tagged with ``O'', it means the word is not an entity. The CoNLL-2003 English dataset uses the IOB tagging method. BIO tagging is a derivative method of IOB tagging. BIO stipulates that all named entities start with ``B'', ``I'' means inside the named entity, and ``O'' means is not a named entity. The labeling method of the dataset used in this paper is BIO tagging. If a word in the corpus is tagged ``B/I-XXX'', ``B/I'' means that the word belongs to the beginning or the inside of the named entity, that is, the word is part of the named entity, and ``XXX'' indicates the type of the named entity.

\subsection{Word Representation}
In recent years, word embedding has transitioned from non-contextualization to contextualization. Typical non-contextualized word embedding tools include GloVe\cite{pennington2014glove}, word2vec\cite{mikolov2013distributed}, fastText\cite{bojanowski2017enriching}, etc. CWR (Contextualized Word Representations) has already dealt with many entity-related tasks. BERT\cite{devlin2019bert} is the most typical example of CWR, BERT is based on bidirectional Transformer and uses the MLM method for training. The so-called MLM method is to randomly mask the words in the text before training, and predict the masked words after training. Other models based on CWR include RoBERTa\cite{liu2019roberta}, Albert\cite{lan2019albert}, Xlnet\cite{yang2019xlnet}, SpanBert\cite{joshi2020spanbert}, BART\cite{zhang2020pegasus}, etc.

BERT\cite{devlin2019bert} is a large-scale pre-trained language representation model proposed by Google back in 2018. BERT is internally a bi-directional Transformer \cite{vaswani2017attention} that uses self-attention to capture relations between words. The input vector of the BERT is formed by summing up the three Embedding of Token Embedding, Segment Embedding and Position Embedding, and the output is formed by concatenating the hidden layer vectors of the bi-directional Transformer.

The innovation of BERT lies in the pre-training method, which captures the vector representation of words and sentences using two methods: Masked Language Model (MLM) and Next Sentence Prediction (NSP) respectively. Of the randomly masked tokens, 10$\%$ are replaced with other words, 10$\%$ are not replaced, and the remaining 80$\%$ are replaced with "[MASK]". The main purpose of NSP is to make the model pay more attention to the connection between two sentences, so it can be used for tasks such as intelligent question answering and 
natural language inference. BERT loaded with pre-trained models can be applied to downstream tasks, such as named entity recognition.

\subsection{Contrastive Learning}
Many researchers believe that the essence of deep learning is to do two things: Representation Learning and Inductive Bias Learning. Representation learning do not necessarily focus on every detail of the sample, but only on the features learned by the sample that can be distinguished from other samples. In terms of word ambiguity, we only need to ensure that words with the same entity type are relatively similar, and words with different entity types are relatively dissimilar. In this regard, it leads to contrastive learning. The main idea of contrastive learning is to encode the original sample data through the data augmentation method, and then use representation learning to determine whether the augmented data pair comes from the same sample. The main goal of contrastive learning is to learn a mapping function $f$, the sample is encoded by the mapping function $f$ as $f(x)$ so that:
\begin{equation}
    s\left(f\left(x\right),f\left(x^+\right)\right)\gg\ s\left(f\left(x\right),f\left(x^-\right)\right)
\end{equation}

$x$ represents the common sample, and $x^+$ represents a positive sample of $x$, and $x^-$ can be regarded as a negative sample of $x$, $s\left(\cdot, \cdot \right)$ is a function to measure the similarity of samples. A typical similarity function is to find the dot product of the sample representation vectors. The loss function used in contrastive learning is InfoNCE\cite{oord2018representation}.

Minimizing InfoNCE can maximize the lower bound of Mutual Information of $f\left(x\right)$ and $f\left(x^+\right)$, making the representations of a positive sample pair closer than those of a negative pair in the comparison example. In contrastive learning, how to construct positive and negative sample pairs is an important problem.

\begin{figure}[htbp]
\centering
\includegraphics[scale=0.28]{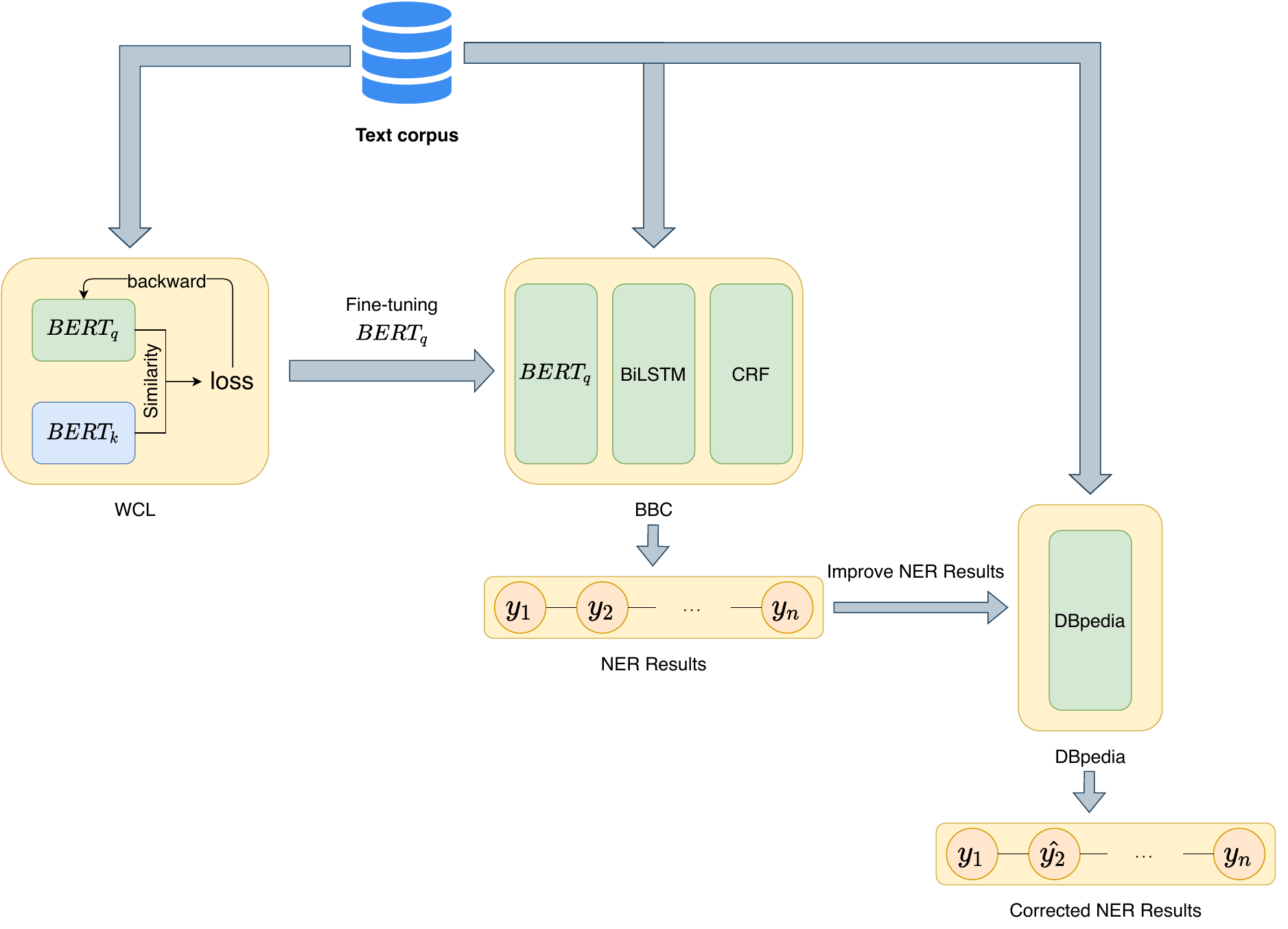}
\caption{The flow chart of WCL-BBCD.}
\label{WCL-BBCD}
\end{figure}

In the field of Computer Vision, positive pairs or negative pairs can be constructed by cropping, rotating, filtering, and graying images. In the field of natural language processing, Wu\cite{wu2020clear}, Meng\cite{meng2021coco} proposed to randomly delete a word in a sentence, rearrange the words in the sentence, or replace some words in the sentence, to construct positive pairs. However, due to the discreteness of words, the above operations are likely to destroy the original meaning of the sentence. In order to keep the original meaning of the sentence, Gillick\cite{gillick2019learning}, Karpukhin\cite{karpukhin2020dense} proposed to extract sentences with similar meanings from different corpora, and use two encoders to encode the extracted sentences. Logeswaran\cite{logeswaran2018efficient} proposed to use two consecutive sentences as a positive pair. Zhang\cite{zhang2020unsupervised} proposed to use the Sentence Embedding and Word Embedding of a sentence as a positive sample pair. Robinson\cite{robinson2020contrastive} proposed to treat samples with different categories as negative samples. Zhou\cite{zhou2020pre} developed a joint pre-training framework for incremental pre-training language models. Patrick\cite{patrick2020support} used generative models to aggregate samples with similar semantics. Gao\cite{gao2021simcse} fed a sample into the encoder twice, using different dropouts to construct positive and negative samples. Su\cite{su2021whitening} applied whitening to improve the isotropy of the learned representation and also to reduce the dimensionality of sentence embedding.

\section{The Proposed Model}
This section describes the framwork and various components of the proposed WCL-BBCD model. WCL-BBCD consists of three parts: WCL model, BBC (\textbf{B}ERT-\textbf{B}iLSTM-\textbf{C}RF) model, and DBpedia. The flow chart of WCL-BBCD is shown in Fig.  \ref{WCL-BBCD}. WCL inputs two sentences with similar semantics into the two BERT to obtain the word embedding corresponding to each word in the two sentences and uses the contrastive loss function to measure the similarity in the vector representation space, and in the training process using gradient descent algorithm to fine-tune BERT parameters. After training, use the fine-tuned BERT combined with the BiLSTM-CRF to perform named entity recognition tasks, and then incorporate the public knowledge graph DBpedia to further improve the recognition accuracy.

\begin{figure}[htbp]
\centering
\includegraphics[scale=0.38]{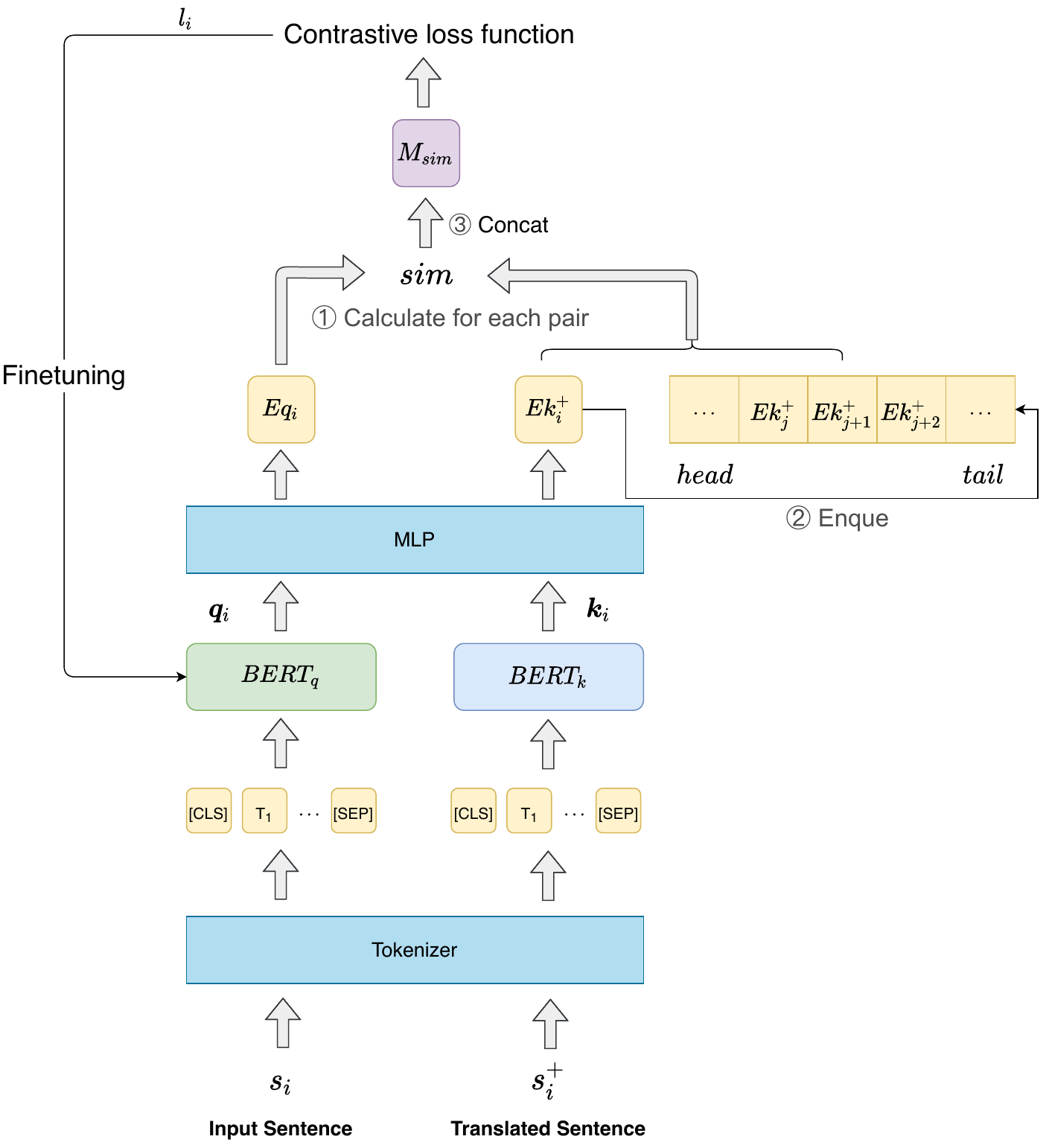}
\caption{Architecture of WCL Model.}
\label{WCL}
\end{figure}

\begin{figure}[htbp]
\centering
\includegraphics[scale=0.32]{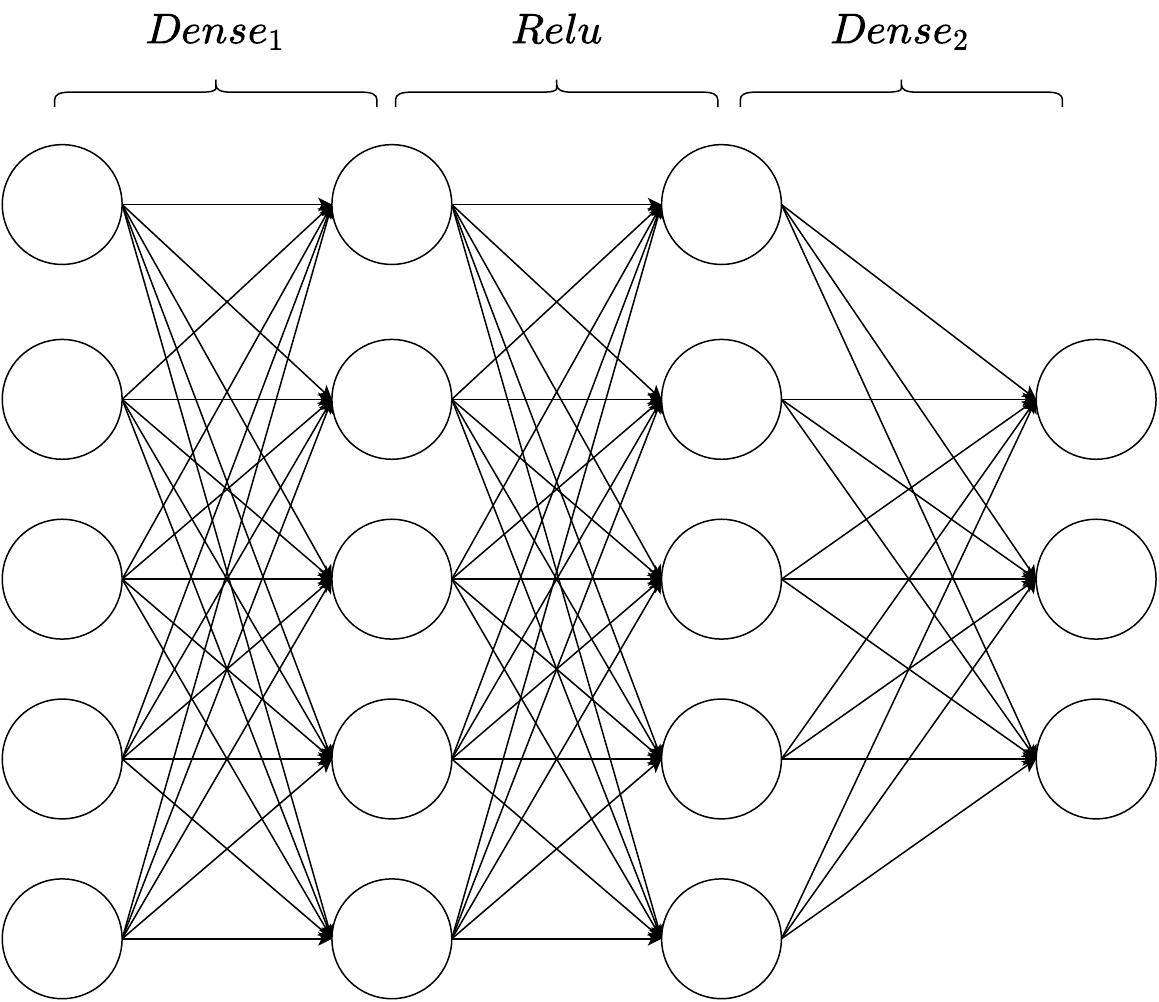}
\caption{Architecture of MLP.}
\label{MLP}
\end{figure}

\subsection{WCL Model}
WCL is proposed to fine-tune BERT in downstream named entity recognition tasks. The core framework of WCL is shown in Fig. \ref{WCL}. The input of WCL is a positive pair of sentences $s_i$ and $s_i^+$ with similar semantics. In order to obtain the positive pair $s_i$ and $s_i^+$, the sentence $s_i$ is translated into sentence $a_i$ in any language, and then the sentence $a_i$ is translated back into the same language as $s_i$. Construct a set of positive and negative sample sentence pairs from the processed sentences. The positive sentence pair set $S^+$ is composed of positive pair of $s_i$ and $s_i^+$, and the negative sample sentence pairs set $S^-$ is composed of $s_i$ and the sentences translated from the rest of the sentences in the corpus. The formal definition of $S^+$ and $S^-$ is given as follows:
\begin{equation}
    \begin{cases}
    S^+=\left\{\left(s_i,s_i^+\right)\right\}_{i=1}^m \\
    S^-=\left\{\left(s_i,s_j^+\right)\right\}_{i=1,j=1}^m,i\neq j
    \end{cases}
\end{equation}
where, $m$ represents the total number of positive and negative sample sentence pairs.

We use two BERT encoders $BERT_q$ and $BERT_k$ to encode $s_i$ and $s_i^+$ to obtain $q_i$ and $k_i^+$ respectively. To better train word embeddings, we add a multi-layer perceptron (MLP) which is composed of the first fully connected layer ($Dense_1$), the Relu activation function, and the second fully connected layer ($Dense_2$) to WCL. $Dense_1$ is transformed into an output vector of the same dimension through a layer of linear change, and the Relu activation function can maintain the convergence rate of the model in a stable state. $Dense_2$ converts the output vector of the Relu activation into an output vector (named generalized representation vector) with a dimension same as the number of predicted entity types. The schematic diagram of the MLP structure is shown in Fig. \ref{MLP}. Taking $q_i$ and $k_i^+$ as the input of MLP, and obtaining the generalized representation vectors $Eq_i$ and $Ek_i^+$ of MLP, the process can be expressed as:
\begin{equation}
    \begin{cases}
    Eq_i=W_2relu(W_1q_i+b_1)+b_2 \\
    Ek_i^+=W_2relu(W_1k_i^++b_1)+b_2
    \end{cases}
\end{equation}
where, $W_1$ and $b_1$ are the weight matrix and bias vector of $Dense_1$, respectively, $W_2$ and $b_2$ are the weight matrix and bias vector of $Dense_2$, respectively.

After obtaining the outputs of the MLP, the model first computes the similarity $r_1^+$ between positive pair of $Eq_i$ and $Ek_i^+$ using the similar function $sim\left(\cdot \right)$. In order to compute the similarities $\left(r_2^-,\ldots, r_N^-\right)$ between negative pairs of $Eq_i$ and $Ek_j^+,j\neq i$ using $sim\left(\cdot \right)$, a queue with a length of $N-1$ is used to store the generalized representation vectors of negative samples of $Eq_i$. At the very beginning, the vectors in the queue are randomly initialized to the standard normal distribution. After calculating the similarities of positive pair and negative pairs for $Eq_i$, then $Ek_i^+$ will enqueue and the earliest vector will dequeue to ensure that there are $N-1$ negative samples in the queue. Then, concatenating the similarity of positive pair ($r_1^+$) and the similarities of negative pairs $\left(r_2^-,\ldots, r_N^-\right)$ into a new vector named $M_{sim}$. The advantage of using queue is that the queue size and batch size can be decoupled, that is, the queue size is no longer limited by the batch size constraint, which solves the problem of requiring a large number of negative samples in SimCLR\cite{chen2020simple}.

In this paper, InfoNCE\cite{oord2018representation} is used as the contrastive loss function to measure the similarity of the samples in the vector representation space. The detailed definition is:
\begin{equation}
    l_i=-\log{\frac{\exp{\left(sim(Eq_i,Ek_i^+)/\tau\right)}}{\sum_{j=1}^{N}\exp{\left(sim(Eq_i,Ek_j^+)/\tau\right)}}}
\end{equation}
where, $l_i$ represents the InfoNCE loss of current mini-batch data (one positive pair and $N-1$ negative pairs of $s_i$), $\tau$ is a hyperparameter, and its function is to adjust the similarity to the input level of InfoNCE. The smaller the value of the loss function, the higher the similarity of the positive sample pair and the lower the similarity of the negative sample pair. Note:
\begin{equation}
    \begin{cases}
    r_i^+=sim(Eq_i,Ek_i^+)\\
    r_j^-=sim(Eq_i,Ek_j^+),i \neq j
    \end{cases}
\end{equation}

\begin{algorithm}[H]
\caption{The training process of WCL.}
\label{alg:wcl}
\begin{algorithmic}[1]
\REQUIRE Sentence pair
\ENSURE Fine-tuned BERT
\STATE $BERT_q.params$=$BERT_k.params$
\WHILE{$epoch_i<epoch_{nums}$}
\FOR{$mini\_batch$ in $dataloader$}
\STATE $Eq_i=mlp\left(BERT_q\left(s_i\right)\right)$
\STATE $Eq_i=normalize(Eq_i)$
\STATE \textbf{with} nograd \textbf{do}
\STATE \quad $Ek_i^+=mlp\left(BERT_k(s_i^+)\right)$
\STATE \quad $Ek_i^+=normalize(Ek_i^+)$
\STATE 
\STATE $pos=sim(Eq_i,Ek_i^+)$
\STATE $neg=sim(Eq_i,queue)$
\STATE $dequeue(queue)$
\STATE $enqueue(queue,Ek_i^+)$
\STATE $M_{sim}=concat\left([pos,neg],dim=-1\right)$
\STATE  
\STATE $\#$ [0] represents the similarity of positive pair is the largest
\STATE $loss=InfoNCE\left(M_{sim},\ [0]\right)$
\STATE $loss.backward$
\ENDFOR
\ENDWHILE
\end{algorithmic}
\end{algorithm}

\begin{figure}[htbp]
\centering
\includegraphics[scale=0.32]{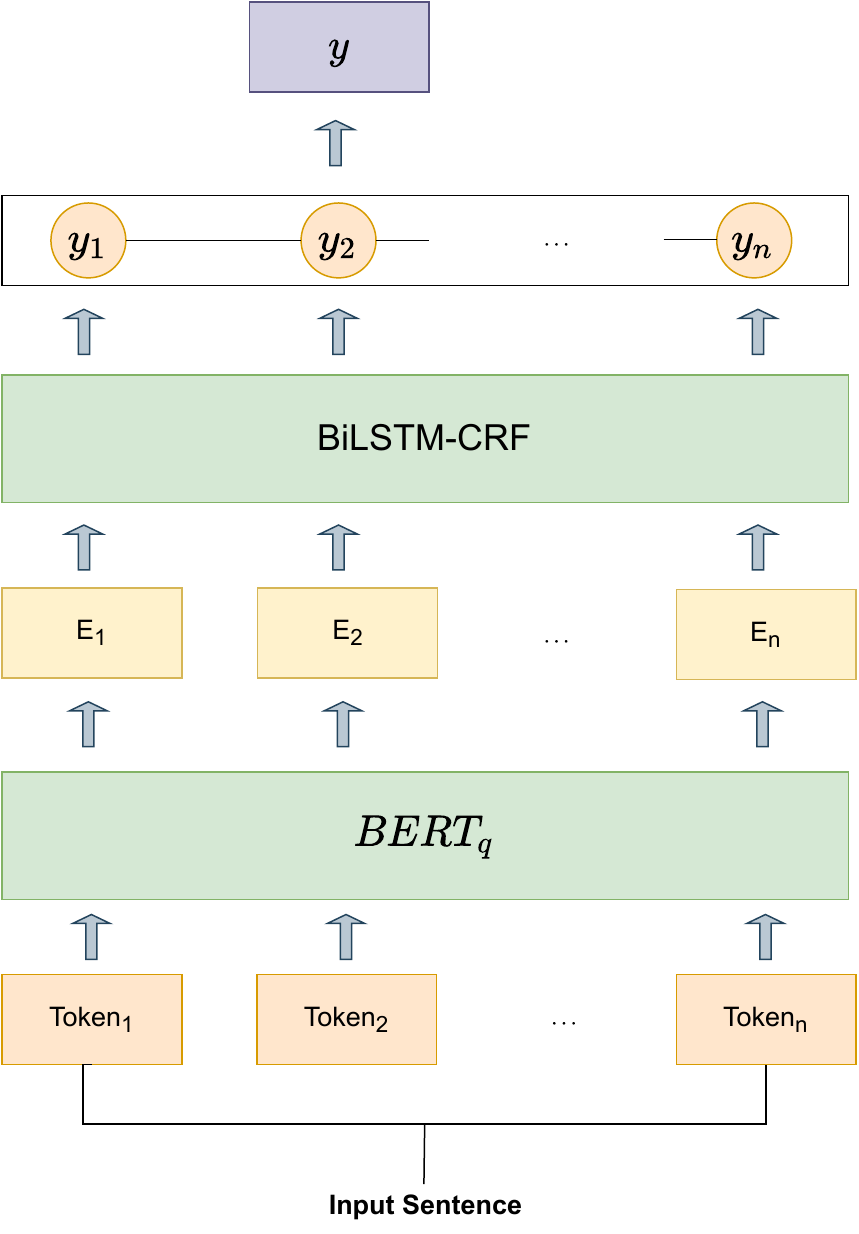}
\caption{Architecture of BBC Model.}
\label{NER}
\end{figure}

In actual computing, the similarity vector $M_{sim}$ is used to record the calculation results of the similarity function. Where $r_i^+$ and $r_j^-$ are aggregated into a similarity vector $M_{sim}$ according to row aggregation. The expression of $M_{sim}$ is:
\begin{equation}
    M_{sim}=\begin{bmatrix}
    r_1^+ \\
    r_2^- \\
    \vdots \\
    r_N^-
    \end{bmatrix}
\end{equation}

According to $M_{sim}$, the above InfoNCE contrastive loss function can be rewritten as:
\begin{equation}
    l_i=-\log{\frac{\exp(r_i^+/\tau)}{sum\left(\exp(M_{sim}/\tau)\right)}}
\end{equation}
$sum(\cdot)$ represents the addition of vector values in rows. The training process of WCL is shown in Algorithm \ref{alg:wcl}.

\subsection{BBC Model}
This section introduces the network layer organization structure of BBC, the architecture of model is shown in Fig. \ref{NER} and contains the following three core components:
\begin{itemize}
    \item \textbf{Fine-tuned BERT layer}: Processing the words in the original text into word embeddings.
    \item \textbf{BiLSTM Layer}: Text is data with contextual dependencies, and the long short-term memory network LSTM\cite{hochreiter1997long} is a variant of RNN, which is also suitable for modeling sequence data. LSTM solves the problem of gradient disappearance and gradient explosion during long sequence training of RNN. In named entity recognition, it is necessary to fully integrate the past and future information, so in this paper, BiLSTM (Bi-directional Long Short-Term Memory) is used as a feature extractor to capture the past and future information.
    \item \textbf{CRF Layer}: There may be some meaningless characters in the output of the BiLSTM layer, and BiLSTM cannot consider the dependency between tags, so we use CRF (Conditional Random Field)\cite{lafferty2001conditional} to reasonably combine context information to further improve the recognition result. CRF is a typical discriminant model, which uses a negative log-likelihood function to train the transition probability between named entity tags so that the recognized entity meets the labeling rules.
\end{itemize}



\subsection{Knowledge Graph}
As a large-scale semantic relationship network, knowledge graph extracts, constructs, and stores knowledge from a large amount of data and serves for various purposes. Knowledge graph describes the real world in the form of a symbolic structure. The knowledge graph is a graph-based data structure, which is composed of graph nodes and edges to form a knowledge network. The graph nodes are entities, and the edges with annotations are entity relationships. At present, there are two mainstream data models for knowledge graphs, namely RDF graphs and attribute graphs. The RDF graph is defined as a set of triples, which is a general representation of the knowledge graph\cite{wang2017knowledge}, and its expression is $(h,r,t)$, where $h$ represents the head entity, and $r$ represents the relationship between entities, $t$ represents the tail entity. For example, triplet $(Yimou\ Zhang, rdf:type, director)$ indicates that $Yimou Zhang$ is a director.

This paper chooses DBpedia\cite{auer2007dbpedia} as the correction module to correct the output by BBC. Its data source comes from Wikipedia entries. Currently, DBpedia contains about 768 types of entities and about 3000 different attributes.

Since the CRF layer can only consider the dependency between tags, but still cannot solve the problem of some entity recognition errors, we add DBpedia to identify the entity category recognition errors caused by word abbreviations. This paper mainly includes two modules in the process of correcting the output results: the phrase retrieval module and the entity modification module.
\subsubsection{Phrase Retrieval Module}
The phrase retrieval module is used to obtain a set of potential entities of the input items in BBC and filter out the potential entities existing in DBpedia. The set of potential entities includes single words and phrases composed of multiple words. The input of the phrase retrieval module is a sentence, and the output is a set of potential entities $PE$ in the sentence corpus. The specific steps of constructing a potential entity set are: 
\begin{itemize}
    \item First, get a word with all uppercase letters, and then search the corpus for the phrases composed of multiple words, of which first letters are the same as the uppercase letters of the word, respectively. For example, the word ``TEC'' is corresponding to the phrase ``The European Commission''.
    \item Find the permutation and combination of all words in the phrase into words and phrases set. For example, the phrase ``The European Commission'' can generate the set \emph{Pe=\{The, European, Commission, The European, European Commission, The European Commission\}}.
    \item Enter each potential entity in the set $Pe$ into DBpedia for retrieval. If the potential entity and the entity type corresponding to the potential entity exist in the public knowledge graph, add the potential entity and entity type to the potential entity set $PE$, for example, $PE=\{The\ European\ Commission: Organization,\ldots\}$.
\end{itemize}

\subsubsection{Entity Modification Module}
The entity modification module is used to receive the potential entity set $PE$ generated by the phrase retrieval module, and the entity type label output by BBC, and then compare the entity type labels output by BBC with the entity types corresponding to the potential entities in the potential entity set $PE$. If they are consistent, there is no need to modify them. If they are inconsistent, BBC output results are modified. The Entity Modification Module can effectively alleviate the problem of entity category recognition errors caused by word abbreviations.

\section{Experiments}
\subsection{Experimental Setup}
In this section, experiments on the CoNLL-2003 English dataset and OntoNotes V5 English dataset will be conducted. For the two datasets, the BIO tagging method is used will be conducted, and the evaluation indicators in the experimental results are calculated by the ``conlleval'' script. There are four types of entities in the English dataset of CoNLL-2003: Person, Location, Organization, and Miscellaneous (MISC). The corpus in the OntoNotes V5 English dataset is composed of 5 parts\cite{pradhan2013towards}: news line (300k), broadcast news (200k), broadcast dialogue (200k), P2.5 data (145k), and network data (200k). There are 18 entity types in the OntoNotes V5 English dataset. This paper uses the official test set for testing. Table \ref{DBpedia-statistics} shows the data statistics of the CoNLL-2003 English dataset and OntoNotes V5 English dataset.

We use Precision, Recall, and Micro F1-score as the evaluation metrics to evaluate the performance of the proposed model.

\begin{table}[htbp]
\caption{Data statistics of CoNLL-2003 English dataset and OntoNotes V5 English dataset}
\label{DBpedia-statistics}
\centering  
\renewcommand{\arraystretch}{1.5}
\begin{tabular}{|c|c|c|c|c|}  
\hline
Dataset & Types & Train & Dev & Test\\ \hline  
\multirow{2}*{CoNLL-2003} & Tokens & 203,621 & 51,362 & 46,435 \\ \cline{2-5}
		~ & Entities & 23,499 & 5,942 & 5,648 \\ \hline
\multirow{2}*{OntoNotes V5} & Tokens & 1,088,503 & 147,724 & 152,728 \\ \cline{2-5}
		~ & Entities & 81,828 & 11,066 & 11,257 \\ \hline
\end{tabular}
\end{table}

The hardware environment of the experiment in this paper is a single-machine with a single-card (GPU), including an Intel(R) Xeon(R) CPU @ 2.20GHz CPU with a memory size of 25GB and an NVIDIA Tesla P100 with a memory size of 16GB.

\subsection{Evaluation Metrics}
At present, there are two commonly used evaluation methods for named entity recognition, which are exact matching evaluation and loose matching evaluation. In this paper, the exact matching evaluation is adopted.

The tagging method in the named entity recognition determines the entity boundary and entity type at the same time. In the exact matching evaluation, only if the entity boundary and entity type are accurately marked at the same time, the current entity recognition result is regarded as correct. Based on the True Positive (TP), False Positive (FP), and False Negative (FN) of the data, Precision, Recall and F1-score of the named entity recognition task can be calculated. In the task of named entity recognition, TP is defined as being able to correctly identify entity boundaries and entity types. FP is defined as identifying an entity but the entity boundary or entity type is judged incorrectly. FN is defined as an entity that should be recognized but not recognized.

According to the definition of Precision: for a given dataset, precision is the ratio of the number of correctly recognized samples to the total number of samples. The calculation method of precision in the named entity recognition task is given by equation:
\begin{equation}
    Precision=\frac{TP}{TP+FP}
\end{equation}

According to the definition of Recall: Recall is used to describe the ratio of the positive cases judged to be true in the classifier to the total proportion. The calculation method of recall in the named entity recognition task is given by equation:
\begin{equation}
    Recall=\frac{TP}{TP+FN}
\end{equation}

According to the definition of F1-score: F1-score is the harmonic average index of precision and recall, and is a comprehensive index that balances the influence of precision and recall. The calculation equation of F1-score can be obtained:
\begin{equation}
    F1=2\times\frac{Precision\times Recall}{Precision+Recall}
\end{equation}

There are two types of F1-score, namely macro-averaged F1-score, and micro-averaged F1-score. Macro-average F1-score is to calculate F1-score for each entity category separately, and then calculate the overall average. Micro-average F1-score is to calculate F1-score for the overall data.

MUC-6\cite{chinchor1995muc} defines a loose match evaluation standard: as long as the words of the predicted entity has an nonempty intersection with the words of the actual entity and the entity type is correct, the entity identification is regarded as correct. The recognition of the entity boundary is not required to judge whether the entity type recognition is correct. In general, loose matching evaluation is less frequently used.

\subsection{Baselines}
In this paper, we use BiLSTM as the basic model. Based on this model, we expand and conduct experiments. For example, add some word embedding components for BiLSTM, and the specific comparison model settings are as follows:
\begin{enumerate}
    \item BiLSTM: A model consists of BiLSTM only.
    \item BiLSTM-CRF: A model consists of BiLSTM and CRF.
    \item CNN-BiLSTM: A model consists of CNN and BiLSTM, CNN is used for character-level encoding.
    \item CNN-BiLSTM-CRF: A model consists of CNN, BiLSTM and CRF.
    \item Glove-BiLSTM: A model consists of Glove and BiLSTM.
    \item Glove-BiLSTM-CRF: A model consists of Glove, BiLSTM and CRF.
    \item Glove-CNN-BiLSTM: A model consists of Glove, CNN, and BiLSTM.
    \item Glove-CNN-BiLSTM-CRF: A model consists of Glove, CNN, BiLSTM and CRF.
    \item BERT-BiLSTM-CRF (BBC): A model consists of BERT, BiLSTM and CRF.
    \item WCL-BBC: The named entity recognition model based on contrastive learning proposed in this paper without Dbpedia.
    \item WCL-BBCD: The complete named entity recognition model proposed in this paper.
\end{enumerate}

The word embedding of models 1) - 4) using Pytorch Embedding. The word embedding of models 5) - 8) using the pre-trained word embedding of the Glove algorithm. The word embedding of  models 9) - 11) are obtained through training BERT.

\subsection{Experimental Results}
The experiments are carried out in the experimental environment mentioned above. The experimental results on the CoNLL-2003 and OntoNotes V5 test sets are shown in Table \ref{CoNLL-2003 experimental results} and Table \ref{OntoNotes V5 experimental results}. In all the tables, improvement = (bold number - underlined number) / underlined number.

Glove in the experiment refers to the use of pre-trained word embedding, the dimension of the word embedding is 100. The BERT used in the experiments is $bert-base-cased$, and the case of words needs to be distinguished in the task of named entity recognition. As shown in Table \ref{CoNLL-2003 experimental results} and Table \ref{OntoNotes V5 experimental results}, WCL-BBCD proposed in this paper performs the best among all models on both CoNLL-2003 and OntoNotes V5 test sets. Compared with BBC, the F1-score has been improved 1.59$\%$ on the CoNLL-2003 dataset, and 0.80$\%$ on the OntoNotes V5 dataset, respectively.

\begin{table}[htbp]
\caption{Experimental results on CoNLL-2003 dataset}
\label{CoNLL-2003 experimental results}
\centering  
\renewcommand{\arraystretch}{1.5}
\begin{tabular}{|c|c|c|c|}  
\hline
\multirow{2}*{Model} & \multicolumn{3}{c|}{CoNLL-2003}\\ \cline{2-4}
~ & P & R & F1\\ \hline 
BiLSTM	&81.90&73.02&77.20\\ \hline
BiLSTM-CRF	&84.78&73.07&78.49\\ \hline
CNN-BiLSTM	&75.72&80.83&78.20\\ \hline
CNN-BiLSTM-CRF	&80.49&81.47&80.98\\ \hline
Glove-BiLSTM	&88.19&88.30&88.24\\ \hline
Glove-BiLSTM-CRF &89.71&87.68&88.68\\ \hline
Glove-CNN-BiLSTM	&89.03&90.44&89.73\\ \hline
Glove-CNN-BiLSTM-CRF	&90.56&90.86&90.71\\ \hline
BBC &\underline{91.24}&\underline{91.52}&\underline{91.38}\\ \hline
WCL-BBC &91.47&92.07&91.77\\ \hline
\textbf{WCL-BBCD} &\textbf{91.88}&\textbf{93.81}&\textbf{92.83}\\ \hline
Improvement &0.70$\%$&2.50$\%$&1.59$\%$\\ \hline
\end{tabular}
\end{table}

\begin{table}[htbp]
\caption{Experimental results on OntoNotes V5 dataset}
\label{OntoNotes V5 experimental results}
\centering  
\renewcommand{\arraystretch}{1.5}
\begin{tabular}{|c|c|c|c|}  
\hline
\multirow{2}*{Model} &\multicolumn{3}{c|}{OntoNotes V5} \\ \cline{2-4}
~ & P & R & F1\\ \hline 
BiLSTM	&80.30&76.79&78.51\\ \hline
BiLSTM-CRF	&83.63&79.44&81.48\\ \hline
CNN-BiLSTM	&75.25&80.58&77.82\\ \hline
CNN-BiLSTM-CRF	&82.50&80.88&81.68\\ \hline
Glove-BiLSTM	&81.28&83.53&82.39\\ \hline
Glove-BiLSTM-CRF &86.20&85.28&85.74\\ \hline
Glove-CNN-BiLSTM	&81.88&85.20&83.51\\ \hline
Glove-CNN-BiLSTM-CRF	&85.36&85.59&85.48\\ \hline
BBC &\underline{88.16}&\underline{88.82}&\underline{88.49}\\ \hline
WCL-BBC &88.68&89.38&89.03\\ \hline
\textbf{WCL-BBCD} &\textbf{88.78}&\textbf{89.62}&\textbf{89.20} \\ \hline
Improvement &0.70$\%$&0.90$\%$&0.80$\%$\\ \hline
\end{tabular}
\end{table}

\begin{table*}[htbp]
\caption{Experimental data table on the CoNLL-2003 dataset and OntoNotes V5 dataset after adding DBpedia}
\label{CoNLL-2003-DBpedia}
\centering  
\renewcommand{\arraystretch}{1.5}
\begin{tabular}{|c|c|c|c|c|c|c|c|c|c|c|c|c|}  
\hline
\multirow{3}*{Model} & \multicolumn{6}{c|}{CoNLL-2003} & \multicolumn{6}{c|}{OntoNotes V5}\\
\cline{2-13}
~ & \multicolumn{3}{c|}{Without DBpedia} & \multicolumn{3}{c|}{With DBpedia} & \multicolumn{3}{c|}{Without DBpedia} & \multicolumn{3}{c|}{With DBpedia}\\
\cline{2-13}
~ & P & R & F1 & P & R & F1 & P & R & F1 & P & R & F1\\ \hline
BiLSTM	&81.90	&73.02	&77.20	&83.39	&74.76	&78.84 &80.30	&76.79	&78.51	&81.63	&77.56	&79.55\\ \hline
BiLSTM-CRF	&84.78	&73.07	&78.49	&85.57	&74.60	&79.71 &83.63	&79.44	&81.48	&83.90	&80.10	&81.96\\ \hline
CNN-BiLSTM	&75.72&80.83&78.20&76.86&81.89&79.29&75.25&80.58&77.82&76.33&80.65&78.43\\ \hline
CNN-BiLSTM-CRF	&80.49&81.47&80.98&81.14&82.53&81.83&82.50&80.88&81.68&82.73&81.51&82.11\\ \hline
Glove-BiLSTM	&88.19&88.30&88.24&89.34&89.52&89.43&81.28&83.53&82.39&83.17&84.44&83.80\\ \hline
Glove-BiLSTM-CRF	&89.71&87.68&88.68&90.42&88.88&89.64&86.20&85.28&85.74&86.53&85.88&86.20\\ \hline
Glove-CNN-BiLSTM	&89.03&90.44&89.73&89.92&91.35&90.63&81.88&85.20&83.51&83.70&85.97&84.82\\ \hline
Glove-CNN-BiLSTM-CRF &90.56&90.86&90.71&91.14&91.74&91.44&85.36&85.59&85.48&85.62&86.26&85.94\\ \hline
BBC	&91.24&91.52&91.38&\textbf{91.90}&92.36&92.13&88.16&88.82&88.49&88.53&89.28&88.90\\ \hline
\textbf{WCL-BBC} &\underline{91.47}&\underline{92.07}&\underline{91.77}&91.88&\textbf{93.81}&\textbf{92.83}&\underline{88.68}&\underline{89.38}&\underline{89.03}&\textbf{88.78}&\textbf{89.62}&\textbf{89.20}\\ \hline
Improvement(WCL-BBC, WCL-BBCD) &-&-&-&0.45$\%$&1.89$\%$&1.16$\%$&-&-&-&0.11$\%$&0.27$\%$&1.19$\%$\\ \hline
\end{tabular}
\end{table*}

\begin{figure}[htbp]
\centering
\includegraphics[scale=0.35]{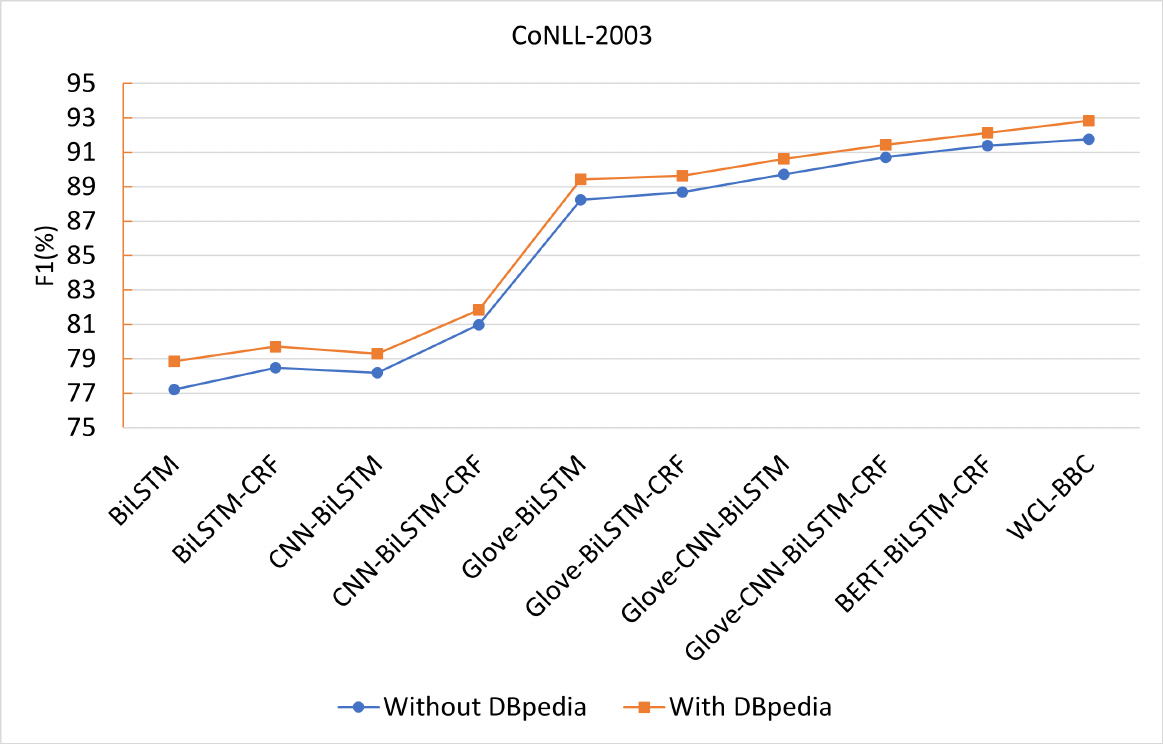}
\caption{F1-score of ablation experiment on the CoNLL-2003 dataset.}
\label{CoNLL-2003-DBpedia-Ablation}
\end{figure}

\begin{figure}[htbp]
\centering
\includegraphics[scale=0.35]{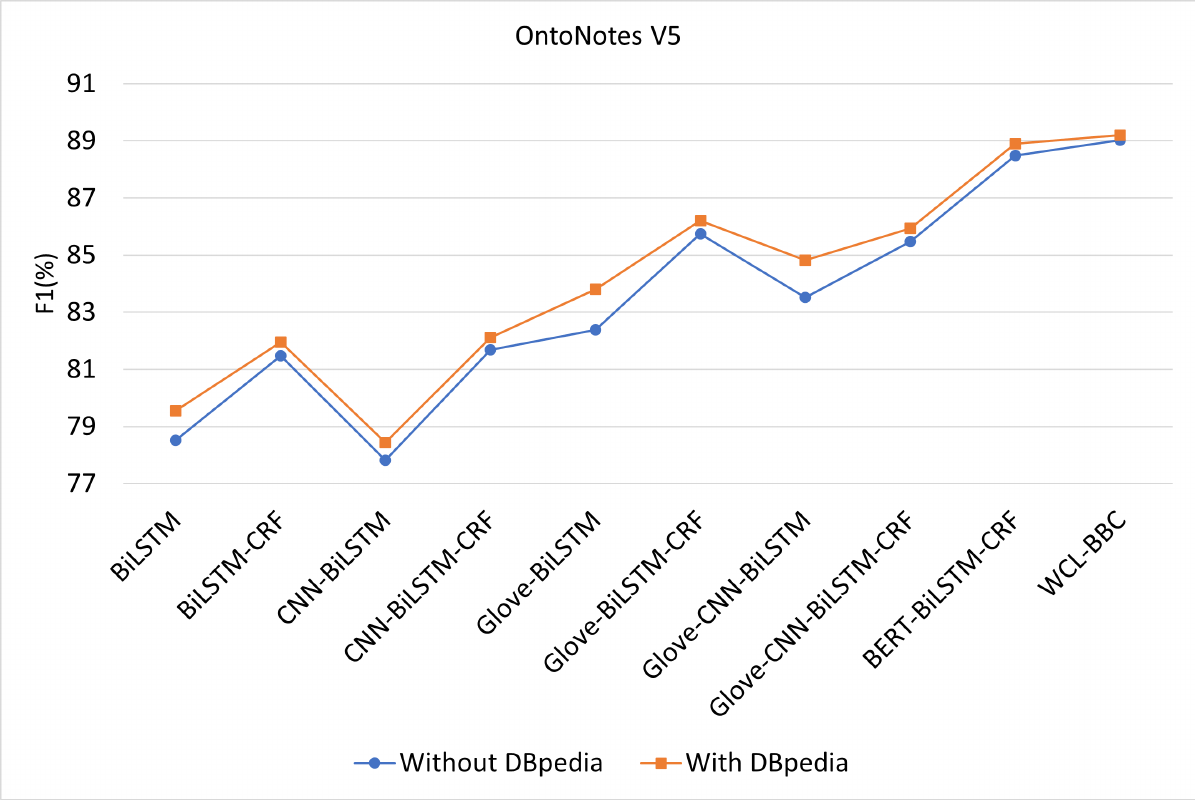}
\caption{F1-score of ablation experiment on the OntoNotes V5 dataset.}
\label{OntoNotes V5-DBpedia-Ablation}
\end{figure}

To verify the effectiveness of the contrastive learning model in named entity recognition, we also conduct ablation experiments, that is, only WCL is added to BBC (WCL-BBC). Since the WCL model needs a Transformer-based embedding encoder during the training process, and the BiLSTM basic model is not based on Transformer, it is only compared with BBC. The ablation experiment demonstrated the effectiveness of contrastive learning in the task of named entity recognition. The analysis is as follows:
\begin{enumerate}
    \item Word embedding plays a vital role in named entity recognition. Compared to using an initial randomly generated word embedding, models that incorporate word embedding components such as Glove and BERT have a greater improvement in prediction. 
    \item WCL improves word representation by learning from a large number of positive sentence pairs and negative sentence pairs during training and uses the queue to decouple it from the batch size, that is, the queue size is no longer limited by the batch size. It solves the problem of a large number of negative mini-batch vector in contrastive learning.
    \item WCL calculates the similarity matrix by training a large number of positive sample pairs and negative sample pairs and combining similarity functions. Words with similar definitions will become more similar after training, and the relevance of words with low correlation will decrease after training, that is, the embedding encoder trained by WCL gets the Token Embedding of a word, it will become more in line with its interpretation in the sentence. Compared with BBC, the F1-score has been improved 0.43$\%$ on the CoNLL-2003 dataset, and 0.61$\%$ on the OntoNotes V5 dataset, respectively.
    \item Fig. \ref{CoNLL-2003-DBpedia-Ablation} and Fig. \ref{OntoNotes V5-DBpedia-Ablation} show the results of the ablation experiments conducted without or with DBpedia. As can be seen from the figures, there is a certain improvement in F1-score for each model after adding DBpedia. The performance improvement of WCL-BBCD over WCL-BBC on the CoNLL-2003 dataset is larger than that on the OntoNotes V5 dataset (the former is 1.16$\%$, and the latter is 0.19$\%$). The reason is that the entity types in the CoNLL-2003 dataset are coarse-grained divisions, while the entity types in the OntoNotes V5 dataset are fine-grained divisions. All the entity types except MISC in the CoNLL-2003 dataset can be retrieved in DBpedia, while the entity types in the OntoNotes V5 dataset, such as Date, Time, Money, and other entity types, cannot be retrieved in DBpedia.
\end{enumerate}

To verify the effectiveness of adding prior knowledge to the named entity recognition task, we only combine DBpedia as a fine-grained comparative test based on BiLSTM-CRF. Table \ref{CoNLL-2003-DBpedia} shows the experimental results on CoNLL-2003 and OntoNotes V5.

As shown in Table \ref{CoNLL-2003-DBpedia}, it can be seen that DBpedia has an improvement effect on each model, with an average improvement of about 1.23$\%$. On the OntoNotes V5 dataset, the improvement effect is 0.82$\%$ on average. It can demonstrate the effectiveness of introducing prior knowledge from the knowledge graph into the named entity recognition tasks.

Combining the above experiments, the performance of WCL-BBCD proposed in this paper is better than other models cited in this paper.

\section{Conclusion}
In this paper, we propose WCL-BBCD for named entity recognition, which incorporates contrastive learning to the field of named entity recognition for the first time. The model improves the embedding encoder to better obtain the word embedding representation. In addition, we propose to integrate knowledge graph into the task of named entity recognition, to correct the words that identify the wrong entity type, and at the same time to extend the recognized entity to the knowledge graph, which plays a complementary role.

Results of experiments conducted on the CoNLL-2003 dataset and OntoNotes V5 dataset demonstrate that the model proposed in this paper achieved the best performance in the three evaluation metrics in comparison with the baseline models. In future work, we will further explore the application of contrastive learning in the field of named entity recognition. In addition, the research on word embedding can be one step further. Combining prior knowledge such as knowledge graphs with word embedding representation is one of our future research directions. The task of named entity recognition can also expand the entity of the knowledge graph, there is a positive interaction between the named entity recognition task and the expansion of the knowledge graph.

\section*{Acknowledgments}
The authors are grateful to the reviewers for their valuable comments and suggests. This work was supported in part by the Key Research and Development Program of Zhejiang Province under Grant 2019C03134, in part by the National Natural Science Foundation of China under Grant 61972358, and in part by the National Key Research and Development Program of China under Grant 2019YFB2102100.

\bibliographystyle{IEEEtran}
\bibliography{reference}

\begin{thebibliography}{10}
\providecommand{\url}[1]{#1}
\csname url@samestyle\endcsname
\providecommand{\newblock}{\relax}
\providecommand{\bibinfo}[2]{#2}
\providecommand{\BIBentrySTDinterwordspacing}{\spaceskip=0pt\relax}
\providecommand{\BIBentryALTinterwordstretchfactor}{4}
\providecommand{\BIBentryALTinterwordspacing}{\spaceskip=\fontdimen2\font plus
\BIBentryALTinterwordstretchfactor\fontdimen3\font minus
  \fontdimen4\font\relax}
\providecommand{\BIBforeignlanguage}[2]{{%
\expandafter\ifx\csname l@#1\endcsname\relax
\typeout{** WARNING: IEEEtran.bst: No hyphenation pattern has been}%
\typeout{** loaded for the language `#1'. Using the pattern for}%
\typeout{** the default language instead.}%
\else
\language=\csname l@#1\endcsname
\fi
#2}}
\providecommand{\BIBdecl}{\relax}
\BIBdecl

\bibitem{zhou2021hybrid}
R.~Zhou, C.~Liu, J.~Wan, Q.~Fan, Y.~Ren, J.~Zhang, and N.~Xiong, ``A hybrid
  neural network architecture to predict online advertising click-through rate
  behaviors in social networks,'' \emph{IEEE Transactions on Network Science
  and Engineering}, vol.~8, no.~4, pp. 3061--3072, 2021.

\bibitem{he2011using}
R.~He, N.~Xiong, L.~T. Yang, and J.~H. Park, ``Using multi-modal semantic
  association rules to fuse keywords and visual features automatically for web
  image retrieval,'' \emph{Information Fusion}, vol.~12, no.~3, pp. 223--230,
  2011.

\bibitem{conlon2015terrorism}
S.~J. Conlon, A.~S. Abrahams, and L.~L. Simmons, ``Terrorism information
  extraction from online reports,'' \emph{Journal of Computer Information
  Systems}, vol.~55, no.~3, pp. 20--28, 2015.

\bibitem{atkinson2009automated}
M.~Atkinson, J.~Piskorski, H.~Tanev, E.~van~der Goot, R.~Yangarber, and
  V.~Zavarella, ``Automated event extraction in the domain of border
  security,'' in \emph{International Conference on User Centric Media}.\hskip
  1em plus 0.5em minus 0.4em\relax Springer, 2009, pp. 321--326.

\bibitem{vanegas2015overview}
J.~A. Vanegas, S.~Matos, F.~Gonz{\'a}lez, and J.~L. Oliveira, ``An overview of
  biomolecular event extraction from scientific documents,''
  \emph{Computational and mathematical methods in medicine}, vol. 2015, 2015.

\bibitem{giorgi2019end}
J.~Giorgi, X.~Wang, N.~Sahar, W.~Y. Shin, G.~D. Bader, and B.~Wang,
  ``End-to-end named entity recognition and relation extraction using
  pre-trained language models,'' \emph{arXiv preprint arXiv:1912.13415}, 2019.

\bibitem{li2004svm}
Y.~Li, K.~Bontcheva, and H.~Cunningham, ``Svm based learning system for
  information extraction,'' in \emph{International Workshop on Deterministic
  and Statistical Methods in Machine Learning}.\hskip 1em plus 0.5em minus
  0.4em\relax Springer, 2004, pp. 319--339.

\bibitem{schapire2013explaining}
R.~E. Schapire, ``Explaining adaboost,'' in \emph{Empirical inference}.\hskip
  1em plus 0.5em minus 0.4em\relax Springer, 2013, pp. 37--52.

\bibitem{devlin2019bert}
J.~Devlin, M.-W. Chang, K.~Lee, and K.~Toutanova, ``Bert: Pre-training of deep
  bidirectional transformers for language understanding,'' in \emph{Proceedings
  of the 2019 Conference of the North American Chapter of the Association for
  Computational Linguistics: Human Language Technologies, Volume 1 (Long and
  Short Papers)}, 2019, pp. 4171--4186.

\bibitem{liu2019roberta}
Y.~Liu, M.~Ott, N.~Goyal, J.~Du, M.~Joshi, D.~Chen, O.~Levy, M.~Lewis,
  L.~Zettlemoyer, and V.~Stoyanov, ``Roberta: A robustly optimized bert
  pretraining approach,'' 2019.

\bibitem{lan2019albert}
Z.~Lan, M.~Chen, S.~Goodman, K.~Gimpel, P.~Sharma, and R.~Soricut, ``Albert: A
  lite bert for self-supervised learning of language representations,'' in
  \emph{International Conference on Learning Representations}, 2019.

\bibitem{yang2019xlnet}
Z.~Yang, Z.~Dai, Y.~Yang, J.~Carbonell, R.~R. Salakhutdinov, and Q.~V. Le,
  ``Xlnet: Generalized autoregressive pretraining for language understanding,''
  \emph{Advances in Neural Information Processing Systems}, vol.~32, pp.
  5753--5763, 2019.

\bibitem{vaswani2017attention}
A.~Vaswani, N.~Shazeer, N.~Parmar, J.~Uszkoreit, L.~Jones, A.~N. Gomez,
  {\L}.~Kaiser, and I.~Polosukhin, ``Attention is all you need,'' in
  \emph{Proceedings of the 31st International Conference on Neural Information
  Processing Systems}, 2017, pp. 6000--6010.

\bibitem{he2020momentum}
K.~He, H.~Fan, Y.~Wu, S.~Xie, and R.~Girshick, ``Momentum contrast for
  unsupervised visual representation learning,'' in \emph{2020 IEEE/CVF
  Conference on Computer Vision and Pattern Recognition (CVPR)}.\hskip 1em plus
  0.5em minus 0.4em\relax IEEE Computer Society, 2020, pp. 9726--9735.

\bibitem{chen2020simple}
T.~Chen, S.~Kornblith, M.~Norouzi, and G.~Hinton, ``A simple framework for
  contrastive learning of visual representations,'' in \emph{International
  Conference on Machine Learning}.\hskip 1em plus 0.5em minus 0.4em\relax PMLR,
  2020, pp. 1597--1607.

\bibitem{sang2003introduction}
E.~T.~K. Sang and F.~De~Meulder, ``Introduction to the conll-2003 shared task:
  Language-independent named entity recognition,'' in \emph{Proceedings of the
  Seventh Conference on Natural Language Learning at HLT-NAACL 2003}, 2003, pp.
  142--147.

\bibitem{pradhan2013towards}
S.~Pradhan, A.~Moschitti, N.~Xue, H.~T. Ng, A.~Bj{\"o}rkelund, O.~Uryupina,
  Y.~Zhang, and Z.~Zhong, ``Towards robust linguistic analysis using
  ontonotes,'' in \emph{Proceedings of the Seventeenth Conference on
  Computational Natural Language Learning}, 2013, pp. 143--152.

\bibitem{jin2018improving}
S.~Jin, H.~Jang, and W.~Kim, ``Improving bidirectional lstm-crf model of
  sequence tagging by using ontology knowledge based feature,'' \emph{Journal
  of intelligence and information systems}, vol.~24, no.~1, pp. 253--266, 2018.

\bibitem{ma2016end}
X.~Ma and E.~Hovy, ``End-to-end sequence labeling via bi-directional
  lstm-cnns-crf,'' in \emph{Proceedings of the 54th Annual Meeting of the
  Association for Computational Linguistics (Volume 1: Long Papers)}, 2016, pp.
  1064--1074.

\bibitem{pennington2014glove}
J.~Pennington, R.~Socher, and C.~D. Manning, ``Glove: Global vectors for word
  representation,'' in \emph{Proceedings of the 2014 conference on empirical
  methods in natural language processing (EMNLP)}, 2014, pp. 1532--1543.

\bibitem{mikolov2013distributed}
T.~Mikolov, I.~Sutskever, K.~Chen, G.~S. Corrado, and J.~Dean, ``Distributed
  representations of words and phrases and their compositionality,'' in
  \emph{Advances in neural information processing systems}, 2013, pp.
  3111--3119.

\bibitem{bojanowski2017enriching}
P.~Bojanowski, E.~Grave, A.~Joulin, and T.~Mikolov, ``Enriching word vectors
  with subword information,'' \emph{Transactions of the Association for
  Computational Linguistics}, vol.~5, pp. 135--146, 2017.

\bibitem{joshi2020spanbert}
M.~Joshi, D.~Chen, Y.~Liu, D.~S. Weld, L.~Zettlemoyer, and O.~Levy, ``Spanbert:
  Improving pre-training by representing and predicting spans,''
  \emph{Transactions of the Association for Computational Linguistics}, vol.~8,
  pp. 64--77, 2020.

\bibitem{zhang2020pegasus}
J.~Zhang, Y.~Zhao, M.~Saleh, and P.~Liu, ``Pegasus: Pre-training with extracted
  gap-sentences for abstractive summarization,'' in \emph{International
  Conference on Machine Learning}.\hskip 1em plus 0.5em minus 0.4em\relax PMLR,
  2020, pp. 11\,328--11\,339.

\bibitem{oord2018representation}
A.~v.~d. Oord, Y.~Li, and O.~Vinyals, ``Representation learning with
  contrastive predictive coding,'' \emph{arXiv preprint arXiv:1807.03748},
  2018.

\bibitem{wu2020clear}
Z.~Wu, S.~Wang, J.~Gu, M.~Khabsa, F.~Sun, and H.~Ma, ``Clear: Contrastive
  learning for sentence representation,'' \emph{arXiv preprint
  arXiv:2012.15466}, 2020.

\bibitem{meng2021coco}
Y.~Meng, C.~Xiong, P.~Bajaj, S.~Tiwary, P.~Bennett, J.~Han, and X.~Song,
  ``Coco-lm: Correcting and contrasting text sequences for language model
  pretraining,'' \emph{arXiv preprint arXiv:2102.08473}, 2021.

\bibitem{gillick2019learning}
D.~Gillick, S.~Kulkarni, L.~Lansing, A.~Presta, J.~Baldridge, E.~Ie, and
  D.~Garcia-Olano, ``Learning dense representations for entity retrieval,'' in
  \emph{Proceedings of the 23rd Conference on Computational Natural Language
  Learning (CoNLL)}, 2019, pp. 528--537.

\bibitem{karpukhin2020dense}
V.~Karpukhin, B.~Oguz, S.~Min, P.~Lewis, L.~Wu, S.~Edunov, D.~Chen, and W.-t.
  Yih, ``Dense passage retrieval for open-domain question answering,'' in
  \emph{Proceedings of the 2020 Conference on Empirical Methods in Natural
  Language Processing (EMNLP)}, 2020, pp. 6769--6781.

\bibitem{logeswaran2018efficient}
L.~Logeswaran and H.~Lee, ``An efficient framework for learning sentence
  representations,'' in \emph{International Conference on Learning
  Representations}, 2018.

\bibitem{zhang2020unsupervised}
Y.~Zhang, R.~He, Z.~Liu, K.~H. Lim, and L.~Bing, ``An unsupervised sentence
  embedding method by mutual information maximization,'' in \emph{Proceedings
  of the 2020 Conference on Empirical Methods in Natural Language Processing
  (EMNLP)}, 2020, pp. 1601--1610.

\bibitem{robinson2020contrastive}
J.~D. Robinson, C.-Y. Chuang, S.~Sra, and S.~Jegelka, ``Contrastive learning
  with hard negative samples,'' in \emph{International Conference on Learning
  Representations}, 2020.

\bibitem{zhou2020pre}
W.~Zhou, D.-H. Lee, R.~K. Selvam, S.~Lee, and X.~Ren, ``Pre-training
  text-to-text transformers for concept-centric common sense,'' in
  \emph{International Conference on Learning Representations}, 2020.

\bibitem{patrick2020support}
M.~Patrick, P.-Y. Huang, Y.~Asano, F.~Metze, A.~G. Hauptmann, J.~F. Henriques,
  and A.~Vedaldi, ``Support-set bottlenecks for video-text representation
  learning,'' in \emph{International Conference on Learning Representations},
  2020.

\bibitem{gao2021simcse}
T.~Gao, X.~Yao, and D.~Chen, ``Simcse: Simple contrastive learning of sentence
  embeddings,'' \emph{arXiv preprint arXiv:2104.08821}, 2021.

\bibitem{su2021whitening}
J.~Su, J.~Cao, W.~Liu, and Y.~Ou, ``Whitening sentence representations for
  better semantics and faster retrieval,'' \emph{arXiv preprint
  arXiv:2103.15316}, 2021.

\bibitem{hochreiter1997long}
S.~Hochreiter and J.~Schmidhuber, ``Long short-term memory,'' \emph{Neural
  computation}, vol.~9, no.~8, pp. 1735--1780, 1997.

\bibitem{lafferty2001conditional}
J.~Lafferty, A.~McCallum, and F.~C. Pereira, ``Conditional random fields:
  Probabilistic models for segmenting and labeling sequence data,'' 2001.

\bibitem{wang2017knowledge}
Q.~Wang, Z.~Mao, B.~Wang, and L.~Guo, ``Knowledge graph embedding: A survey of
  approaches and applications,'' \emph{IEEE Transactions on Knowledge and Data
  Engineering}, vol.~29, no.~12, pp. 2724--2743, 2017.

\bibitem{auer2007dbpedia}
S.~Auer, C.~Bizer, G.~Kobilarov, J.~Lehmann, R.~Cyganiak, and Z.~Ives,
  ``Dbpedia: A nucleus for a web of open data,'' in \emph{The semantic
  web}.\hskip 1em plus 0.5em minus 0.4em\relax Springer, 2007, pp. 722--735.

\bibitem{chinchor1995muc}
N.~Chinchor, ``Muc-6 named entity task definition (version 2.1),'' in \emph{6th
  Message Understanding Conference, Columbia, Maryland}, 1995.

\end{thebibliography}

\end{document}